\documentclass{article}
\usepackage{iclr2027_conference,times}

\usepackage{hyperref}
\usepackage{url}

\usepackage{graphicx}
\usepackage{booktabs}
\usepackage{amsmath}
\usepackage{amssymb}
\usepackage{xcolor}
\usepackage{colortbl}
\definecolor{tfteal}{HTML}{17A1D7}
\definecolor{tforange}{HTML}{E8935A}
\colorlet{oursrow}{tfteal!20}
\colorlet{ablrow}{tforange!25}
\usepackage{pifont}
\usepackage{adjustbox}
\usepackage{wrapfig}

\title{Temporal Forcing: 4D Representation Alignment for Vision-Language-Action Models}

\author{%
\textbf{Xingyu Ding$^{1,2}$, Yuzhong Zhao$^{3}$, Chunhai Zhao$^{2}$, Yinghuan Shi$^{1}$, Chaoyang Zhao$^{2\,\dagger}$,}\\
\textbf{Yifan Zhang$^{2,3\,\dagger}$}\\[4pt]
$^{1}$Nanjing University\\
$^{2}$Institute of Automation, Chinese Academy of Sciences\\
$^{3}$University of Chinese Academy of Sciences\\[2pt]
$^{\dagger}$Corresponding authors
}

\iclrfinalcopy
\begin{document}

\maketitle
\lhead{Preprint}

\begin{abstract}
Long-horizon robotic manipulation requires vision-language-action (VLA) models to track scene states and their evolution beyond the current observation.
However, simply conditioning policies on observation history does not guarantee that the history is effectively utilized: action supervision constrains what the policy should do, but only indirectly constrains what its history representations should retain.
To resolve this, we present Temporal Forcing, a 4D representation alignment framework that explicitly supervises latent temporal states and their transitions.
Specifically, we first introduce a history pathway that compresses past observations into compact latent tokens. We then align these tokens and current-frame features with geometric targets from a pretrained 4D foundation model, providing direct supervision at both the state and transition levels.
The 4D foundation model and alignment heads are used only for training-time supervision.
Temporal Forcing improves average success from 96.6\% to 98.8\% on LIBERO, with the largest gain on LIBERO-Long (93.8\% to 97.2\%), and from 53.5\% to 62.8\% across twelve RoboTwin 2.0 tasks.
Furthermore, Temporal Forcing increases full-task success from 20.0\% to 43.3\% on a physical multi-stage hidden-placement task.
Controlled experiments show that 4D representation alignment is crucial for making observation history beneficial to the model.
Code will be publicly available.
\end{abstract}

\section{Introduction}
\label{sec:intro}

\begin{figure}[t]
\centering
\includegraphics[width=\linewidth]{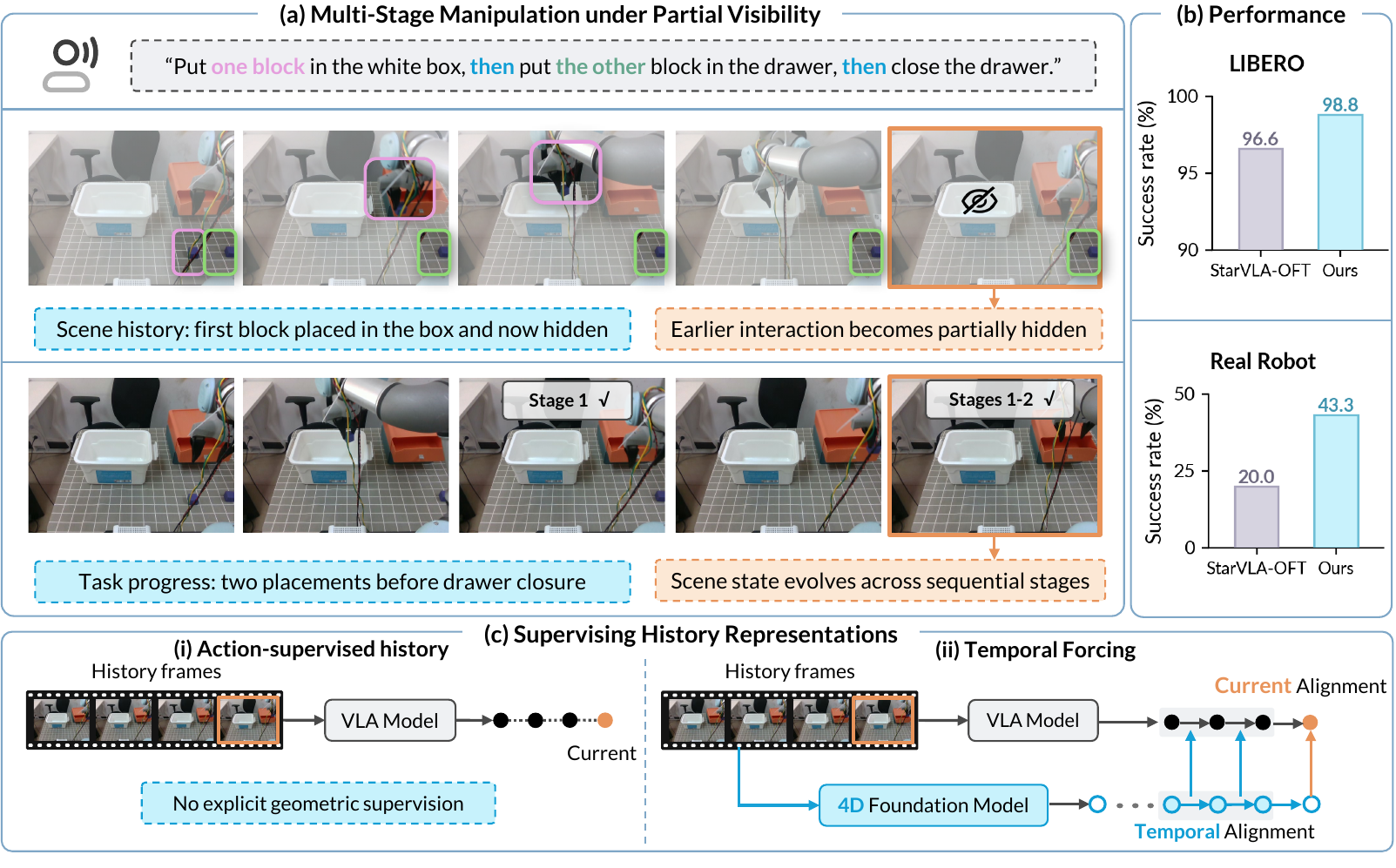}
\caption{
\textbf{Motivation and results of Temporal Forcing.}
(a) A physical multi-stage manipulation task with object occlusion and sequential execution.
(b) Average success on LIBERO and full-task success on the real-robot task.
(c) Temporal Forcing explicitly supervises latent temporal states and their transitions with causal 4D geometric targets, with complementary current-frame alignment.}
\label{fig:teaser}
\end{figure}

Vision-language-action (VLA) models map language instructions and visual observations to robot actions~\citep{rt2,openvla,octo,pi0}.
Long-horizon manipulation, however, requires a history representation that tracks both physical scene evolution and task progress.
For example, as shown in Fig.~\ref{fig:teaser}(a), once one block is placed into an opaque box and becomes invisible, subsequent execution proceeds without full visibility of the earlier interaction. Later, deciding whether to close the drawer requires information about whether both placement stages have been completed, which may no longer be recoverable from the current observation alone. Such sequential decisions can benefit from representations that capture not only the current scene state, but also how it has evolved through recent interactions.

Recent VLA methods incorporate observation history through temporal modules or memory, enabling policies to use past interactions to inform current actions~\citep{hamlet,memoryvla,memoryvlapp}.
However, providing past observations does not directly determine what their compressed representations should retain.
Specifically, action supervision constrains these representations only indirectly through action prediction, without explicitly guiding how they should encode scene states or their changes.
This suggests that the bottleneck is not access to history, but supervision of latent temporal states.

To resolve this, we present Temporal Forcing, a 4D representation alignment framework that explicitly supervises latent temporal states and their transitions (Fig.~\ref{fig:teaser}(c)).
We first introduce a history pathway that compresses a bounded window of past observations into compact history tokens.
To supervise this pathway, we use a pretrained 4D foundation model~\citep{streamvggt} to extract history-conditioned geometric targets from a longer causal context.
We align per-timestep history features with these targets and introduce explicit latent-transition supervision, directly matching inter-timestep changes in the learned history representations to the corresponding changes in causal 4D geometric features.
We further extend this geometric supervision to the VLA backbone through dense current-frame alignment.
The 4D foundation model and alignment heads are used only during training.

Temporal Forcing consistently improves its base model across LIBERO, RoboTwin 2.0, and the real-robot task, with the largest gains in long-horizon or occlusion-heavy settings. Controlled experiments further show that these gains depend on 4D-supervised history rather than history input alone. Our main contributions are as follows:
\begin{itemize}
    \item We identify a supervision bottleneck in history-aware VLAs: action supervision alone does not ensure effective use of observation history, motivating direct supervision of what history representations retain.

    \item We introduce Temporal Forcing, which aligns latent temporal states and their transitions with causal 4D geometric representations, together with a compact History Pathway whose 4D supervision is used only during training.

    \item We validate Temporal Forcing across simulation and real robots, improving LIBERO, RoboTwin 2.0, and multi-stage physical manipulation; ablations and history interventions further confirm that the learned history directly contributes to action prediction.
\end{itemize}

\section{Related Work}
\paragraph{History-augmented VLAs.}
Before the emergence of VLAs, visuomotor policies already incorporated observation history through recurrent models or short observation windows~\citep{robomimic,diffusionpolicy}. Recent VLA methods incorporate longer history through temporal modules or memory. HAMLET~\citep{hamlet} introduces contrastively initialized history tokens, while MemoryVLA~\citep{memoryvla} maintains a perceptual-cognitive memory and MemoryVLA++~\citep{memoryvlapp} further combines memory with future imagination. Other methods encode 3D or 4D history on the input side~\citep{fourdvla,stemvla,vla4d}. These methods provide temporal context to the policy. Temporal Forcing instead directly supervises the latent representations formed from observation history using history-conditioned 4D geometric features.

\paragraph{Training-time representation alignment for VLAs.}
Recent work has also explored supervising VLA representations with features from pretrained models. Aligning intermediate representations with features from pretrained encoders has improved generative models~\citep{repa,geometryforcing}. Spatial Forcing~\citep{spatialforcing} aligns image-token features with VGGT~\citep{vggt}, GLaD~\citep{glad} and ROCKET~\citep{rocket} extend geometric alignment to multiple layers, and QDepth-VLA~\citep{qdepth} predicts quantized depth tokens. Another line introduces explicit 3D inputs such as point clouds or rendered views~\citep{pointvla,geovla,ogvla}, requiring additional geometric inputs or preprocessing at inference. These methods mainly model spatial structure from individual observations. Temporal Forcing applies representation alignment to latent observation history, using temporally conditioned targets to supervise both latent temporal states and their transitions.

\paragraph{Temporal supervision for VLAs.}
Temporal information can also be learned through future prediction or dynamics supervision. World-model-based VLA methods predict or generate future states~\citep{wam4d,dreamvla}, while VLAFlow~\citep{vlaflow} aligns future latent representations and LARA~\citep{lara} aligns VLA representations with latent visual dynamics. StreamVGGT~\citep{streamvggt}, in contrast, causally processes observed sequences and produces temporally consistent geometric representations of the evolving 3D scene. Temporal Forcing instead supervises transitions between latent temporal states, using causal 4D representations as geometric targets rather than predicting future latent states.
\section{Method}
\label{sec:method}

\begin{figure}[t]
\centering
\includegraphics[width=\linewidth]{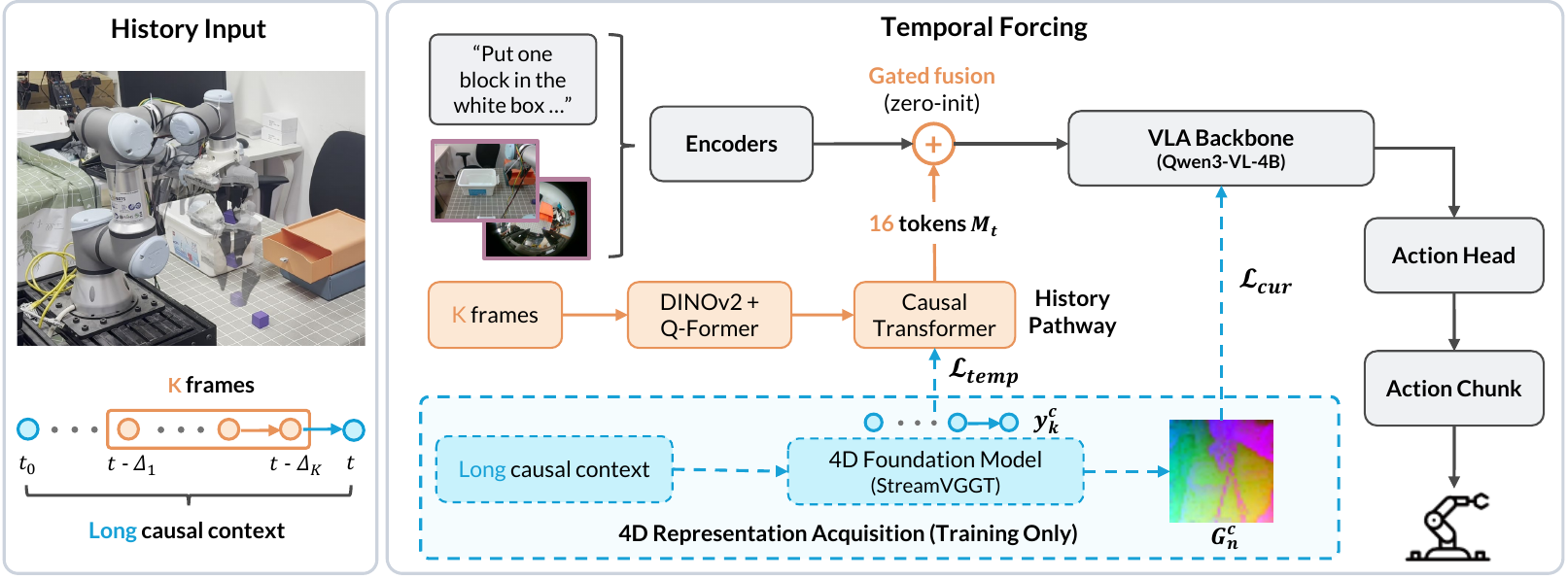}
    \caption{\textbf{Overview of Temporal Forcing.} The History Pathway summarizes $K$ sampled past frames into history tokens that are fused into the VLA model. A streaming 4D foundation model processes a longer causal context and provides geometric targets for temporal alignment ($\mathcal{L}_{\mathrm{temp}}$) and current-frame alignment ($\mathcal{L}_{\mathrm{cur}}$). Blue dashed components are used only during training.}
\label{fig:overview}
\end{figure}

\subsection{Preliminaries}
\label{sec:prelim}
A vanilla VLA model builds on a pretrained vision-language model (VLM) and maps the current observation and a language instruction to robot actions.
Specifically, at each timestep $t$, the model receives visual observations
$\mathbf{o}_t=\{\mathbf{I}_t^c \mid c\in\mathcal{C}\}$ from a set of cameras $\mathcal{C}$
and a language instruction $\ell$.
The VLA backbone jointly encodes the visual observations and the language instruction.
An MLP action head~\citep{openvlaoft} then maps the backbone output to a chunk of $H$ future actions $\hat{\mathbf{A}}_t\in\mathbb{R}^{H\times d_a}$, where $d_a$ denotes the dimensionality of each action.
Given the ground-truth action chunk $\mathbf{A}_t$, the action loss is defined as
\begin{equation}
    \mathcal{L}_{\mathrm{act}}
    =
    \frac{1}{H d_a}
    \left\|
        \mathbf{A}_t-\hat{\mathbf{A}}_t
    \right\|_1 .
    \label{eq:act}
\end{equation}

The vanilla model uses only the current visual observation and does not incorporate observation history.
We therefore introduce a History Pathway to encode past observations and integrate the resulting representations into the VLA backbone.

\subsection{History Pathway}
\label{sec:state}

We design the History Pathway to incorporate observation history with limited inference overhead while preserving the base model’s behavior at initialization. It compresses past observations into a fixed number of tokens and injects them into the current-frame image tokens through gated cross-attention, without extending the backbone sequence.

\paragraph{History window and gist tokens.}
At each timestep $t$, we sample $K$ past frames from each camera at uniformly spaced offsets $\Delta_1 > \cdots > \Delta_K > 0$, ordered from oldest to most recent.
For each sampled frame, a frozen DINOv2 encoder~\citep{dinov2} extracts visual features, which are compressed by a trainable Q-Former~\citep{blip2} into two gist tokens $u_k^c$ for historical timestep $k$ and camera $c$.
Before temporal aggregation, we add time-offset and camera embeddings to each gist token to indicate its temporal position and camera identity.
\paragraph{Temporal summary and gated injection.}
A causal Transformer aggregates the gist tokens at each historical timestep with those from earlier timesteps to produce the History Latent Representation.
To provide the VLA backbone with a compact representation of the history, we use 16 learnable queries to summarize these features into history tokens $M_t$.
The current primary-camera image tokens $x_t$ then attend to $M_t$ through gated cross-attention before the first decoder layer.
This fusion updates the image tokens as $x'_t = x_t + \tanh(\alpha)\,\operatorname{CrossAttn}(x_t, M_t)$, where $\alpha$ is a learnable scalar gate initialized to zero.

\subsection{4D Representation Acquisition}%
\label{sec:representation_acquisition}
We use StreamVGGT~\citep{streamvggt}, a pretrained 4D foundation model, to extract geometric features from observation sequences (Fig.~\ref{fig:overview}).
Compared with the History Pathway, StreamVGGT processes a longer causal context ending at each anchor timestep $t$, reconstructing the observations in a shared coordinate system and using causal attention to condition each frame on earlier observations.
This causal processing yields two complementary geometric targets: frame-level features that describe scene evolution and dense features that preserve the spatial structure of the current frame.
At the $K$ historical timesteps sampled by the History Pathway, we average the tokens of each frame and subtract a fixed per-camera mean $\mu_c$ to obtain the centered feature $\tilde{y}_k^c$, which is layer-normalized to yield the Causal Geometric Feature $y_k^c$.
At the current timestep $t$, we retain the spatial features and layer-normalize each feature vector to obtain the Dense Geometric Features $G_n^c$, where $n$ indexes spatial locations.
Both types of features are precomputed offline at regularly spaced anchor timesteps and remain fixed during training.

\subsection{4D Representation Alignment}
\label{sec:force}
The History Pathway gives the VLA model access to past observations, but the action loss provides only indirect supervision for the history representations.
We therefore introduce 4D representation alignment to guide the model in capturing scene geometry and its changes over time.

\paragraph{Latent state supervision.}
To capture the scene state at each historical timestep, we first align the history representations with the corresponding Causal Geometric Features.
For each historical timestep $k$ and camera $c$, we average the two gist tokens output by the causal Transformer to obtain $\bar{u}_k^c$.
To compare this feature with its geometric target $y_k^c$, we use a learnable projection head $\psi$ to obtain $z_k^c = \psi(\bar{u}_k^c)$.
We minimize their cosine distance using the state loss
\begin{equation}
\mathcal{L}_{\mathrm{state}}
=
\frac{1}{|\mathcal{C}|K}
\sum_{c \in \mathcal{C}}\sum_{k=1}^{K}
\left[1-\cos(z_k^c,y_k^c)\right],
\label{eq:state}
\end{equation}
where $\cos(\cdot,\cdot)$ denotes cosine similarity.
\paragraph{Explicit latent-transition supervision.}
Per-timestep alignment constrains what each history representation encodes, but does not explicitly constrain how the latent state evolves between observations.
We therefore directly supervise latent transitions by matching representation changes between consecutive historical timesteps.
For each camera, we compute $\delta z_k^c = z_{k+1}^c - z_k^c$ and $\delta y_k^c = \tilde{y}_{k+1}^c - \tilde{y}_k^c$, where the target difference is taken on the centered features without layer normalization, and align their directions using the change loss
\begin{equation}
\mathcal{L}_{\mathrm{change}}
=
\frac{1}{|\mathcal{C}|(K-1)}
\sum_{c \in \mathcal{C}}\sum_{k=1}^{K-1}
\left[1-\cos(\delta z_k^c,\delta y_k^c)\right].
\label{eq:change}
\end{equation}
Unlike pointwise feature alignment, $\mathcal{L}_{\mathrm{change}}$ directly constrains the direction of latent-state evolution across adjacent history states.

\paragraph{History readout supervision.} The state and change losses supervise the per-timestep features, while the VLA backbone receives the 16 summarized history tokens.
To supervise these tokens as well, we average them into $\bar{m}$ and align the projected feature $\psi(\bar{m})$ with the geometric targets at the most recent historical timestep using the readout loss
\begin{equation}
\mathcal{L}_{\mathrm{read}}
=
\frac{1}{|\mathcal{C}|}
\sum_{c \in \mathcal{C}}
\left[1-\cos\!\left(\psi(\bar{m}),y_K^c\right)\right].
\label{eq:read}
\end{equation}
This loss provides a direct training signal to the learnable summary queries even when the fusion gate is closed.

We combine the three terms into the \textbf{temporal alignment} objective:
\begin{equation}
\mathcal{L}_{\mathrm{temp}}
=
\frac{1}{6}
\left(
\mathcal{L}_{\mathrm{state}}
+
4\mathcal{L}_{\mathrm{change}}
+
\mathcal{L}_{\mathrm{read}}
\right).
\label{eq:temp}
\end{equation}

\paragraph{Current-frame alignment.}
As a complementary objective, we align the current-frame backbone features with dense geometric features from the same 4D foundation model.
For each camera $c$, we extract the current-frame image features $h_n^c$ from an intermediate backbone layer and project them with a learnable head $\phi$.
We match each projected feature to its corresponding geometric target using
\begin{equation}
\mathcal{L}_{\mathrm{cur}}
=
\frac{1}{|\mathcal{C}|N}
\sum_{c \in \mathcal{C}}\sum_{n=1}^{N}
\left[1-\cos\!\left(\phi(h_n^c),G_n^c\right)\right],
\label{eq:cur}
\end{equation}
where $N$ is the number of spatial locations per camera.
Since StreamVGGT processes the preceding observations when extracting $G_n^c$, the current-frame features receive geometric supervision that incorporates observation history.

\paragraph{Training and inference.}
We train the model with the action loss and the two alignment objectives:
\begin{equation}
\mathcal{L}
=
\mathcal{L}_{\mathrm{act}}
+
\lambda_{\mathrm{cur}}\mathcal{L}_{\mathrm{cur}}
+
\lambda_{\mathrm{temp}}\mathcal{L}_{\mathrm{temp}},
\label{eq:total}
\end{equation}
where $\lambda_{\mathrm{cur}} = \lambda_{\mathrm{temp}} = 0.5$.
The alignment losses are applied at anchor timesteps with precomputed geometric targets; other timesteps use the action loss alone.

At inference, we retain the base VLA model, the History Pathway, and gated fusion; StreamVGGT and the alignment heads are not used.
We cache the per-frame gist tokens before adding time-offset embeddings and reuse them across control steps.

\begin{table}[t]
\caption{\textbf{Performance comparison on LIBERO (success rate, \%).}
The gray-text row evaluates the same Temporal Forcing checkpoint with the injection gate closed at inference.
Best results in each column are in bold.
$^\dagger$ Average recomputed over the four reported suites.}
\label{tab:main}
\begin{center}
\small
\setlength{\tabcolsep}{8pt}
\adjustbox{max width=\textwidth}{
\begin{tabular}{l|cccc|c}
\toprule
  Method & Spatial & Object & Goal & Long & Average \\ \midrule
  \multicolumn{6}{l}{\textit{General VLA}} \\
  TraceVLA \citep{tracevla} & 84.6 & 85.2 & 75.1 & 54.1 & 74.8 \\
  OpenVLA \citep{openvla} & 84.7 & 88.4 & 79.2 & 53.7 & 76.5 \\
  SpatialVLA \citep{spatialvla} & 88.2 & 89.9 & 78.6 & 55.5 & 78.1 \\
  CogACT \citep{cogact} & 97.2 & 98.0 & 90.2 & 88.8 & 93.6 \\
  $\pi_0$ \citep{pi0} & 96.8 & 98.8 & 95.8 & 85.2 & 94.2 \\ \midrule
  \multicolumn{6}{l}{\textit{Framewise 3D alignment}} \\
  GLaD \citep{glad} & 95.0 & 97.4 & 94.4 & 89.4 & 94.1 \\
  Spatial Forcing \citep{spatialforcing} & 99.4 & 99.6 & 98.8 & 96.0 & 98.5 \\ \midrule
  \multicolumn{6}{l}{\textit{History-augmented VLA}} \\
  4D-VLA \citep{fourdvla} & 88.9 & 95.2 & 90.9 & 79.1 & 88.6 \\
  MemoryVLA$^{\dagger}$ \citep{memoryvla} & 98.4 & 98.4 & 96.4 & 93.4 & 96.7 \\
  HAMLET \citep{hamlet} & 99.0 & \textbf{100.0} & \textbf{99.2} & 92.2 & 97.6 \\ \midrule
  \multicolumn{6}{l}{\textit{Same base model (Qwen3-VL-OFT)}} \\
  StarVLA-OFT (base; \citealp{starvla}) & 97.8 & 98.6 & 96.2 & 93.8 & 96.6 \\
  \rowcolor{oursrow} \textbf{Temporal Forcing (Ours)} & \textbf{99.6} & 99.8 & 98.4 & \textbf{97.2} & \textbf{98.8} \\
  \quad w/o 4D alignment (action loss only) & 99.0 & 98.2 & 97.6 & 94.8 & 97.4 \\
  \textcolor{gray}{\quad w/o history at inference (gate closed)} & \textcolor{gray}{98.6} & \textcolor{gray}{97.0} & \textcolor{gray}{99.0} & \textcolor{gray}{94.2} & \textcolor{gray}{97.2} \\
  \bottomrule
\end{tabular}
}
\end{center}
\end{table}
\begin{figure}[t]
\centering
\includegraphics[width=\linewidth]{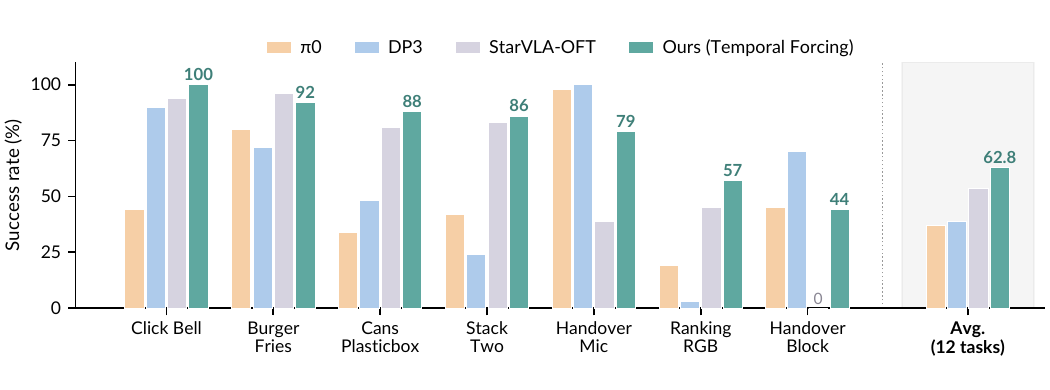}
\caption{\textbf{Performance comparison on RoboTwin 2.0.}
Results include seven representative tasks and the average success rate over all twelve evaluated tasks.
Baselines are $\pi_0$~\citep{pi0}, DP3~\citep{dp3}, and StarVLA-OFT~\citep{starvla}.}
\label{fig:robotwin}
\end{figure}
\section{Experiments}
\label{sec:experiments}

We evaluate Temporal Forcing on LIBERO, RoboTwin~2.0, and a real-robot multi-stage manipulation task. Ablation studies and history-intervention analyses further demonstrate the role of 4D representation alignment in learning to use observation history.

\subsection{Experimental Setup}
\label{sec:setup}

\paragraph{Benchmarks.}
We evaluate on LIBERO~\citep{libero} and RoboTwin~2.0~\citep{robotwin}.
On LIBERO, we train a single model jointly on the Spatial, Object, Goal, and Long suites and evaluate 50 trials per task, totaling 500 trials per suite.
On RoboTwin~2.0, we evaluate twelve bimanual tasks, using 50 clean demonstrations per task for training and 100 trials per task under the easy (clean) setting.
We report success rates and their unweighted averages over tasks.

\paragraph{Implementation Details.}
We use the QwenOFT implementation of StarVLA~\citep{starvla} as the base model, with a Qwen3-VL-4B backbone~\citep{qwen3} and an MLP action head.
For the main LIBERO experiments, we follow the official training recipe for 50k steps with a global batch size of 128 on eight NVIDIA A100 GPUs.
The History Pathway uses $K{=}7$ past frames per camera, spanning 2.8 seconds on LIBERO and 3.7 seconds on RoboTwin~2.0.
History encoding and geometric alignment use the primary and wrist cameras on LIBERO and the head camera on RoboTwin~2.0.
The ablation training protocol is specified in Sec.~\ref{sec:scope}.

\subsection{Main Results}
\label{sec:main_results}

As shown in Table~\ref{tab:main}, Temporal Forcing achieves the highest average success rate of 98.8\% among the compared methods, outperforming its base model by 2.2 percentage points.
The improvement holds across all four suites and is most pronounced on Long, where Temporal Forcing surpasses the base model by 3.4 points (97.2\% vs.\ 93.8\%).
These consistent gains demonstrate the effectiveness of Temporal Forcing for robotic manipulation, particularly on long-horizon tasks.

On RoboTwin~2.0 (Fig.~\ref{fig:robotwin}), Temporal Forcing achieves an average success rate of 62.8\%, outperforming the base model by 9.3 percentage points and improving nine of the twelve tasks.
The largest gains occur on Handover Block and Handover Mic, where success rates increase from 0.0\% to 44.0\% and from 39.0\% to 79.0\%, respectively.
Both tasks involve object occlusions during transfer between the arms and show some of the largest gains from Temporal Forcing.
Fig.~\ref{fig:robotwin} presents seven representative tasks and the average over all twelve tasks.

\subsection{Ablations And History Analysis}
\label{sec:scope}
\begin{table}[t]
\caption{\textbf{Ablation studies on LIBERO (success rate, \%).}
(a) Model components and geometric targets.
(b) Decomposition of temporal supervision.
All variants are trained for 10k steps.
For (b), the tuple in $\mathcal{L}_{\mathrm{temp}}$ denotes
$(w_{\mathrm{state}}, w_{\mathrm{change}}, w_{\mathrm{read}})$ before the
common factor of $1/6$.
Best results in each column are in bold within each panel.}
\label{tab:ablation}
\begin{center}
\setlength{\tabcolsep}{7pt}
\renewcommand{\arraystretch}{1.06}
\newcommand{\abyes}{\ding{51}}
\newcommand{\abno}{\textcolor{black!50}{\ding{55}}}
\adjustbox{max width=\textwidth}{
\begin{tabular}{c cccc cccc c}
\toprule
& \multicolumn{4}{c}{Configuration} & \multicolumn{4}{c}{LIBERO suite} & \\
\cmidrule(lr){2-5}\cmidrule(lr){6-9}
ID & Target & History & $\mathcal{L}_{\mathrm{cur}}$ & $\mathcal{L}_{\mathrm{temp}}$
   & Spatial & Object & Goal & Long & Average \\
\midrule
\multicolumn{10}{l}{\textit{(a) Model components and geometric targets}} \\
\addlinespace[1pt]
1 & \abno & \abno & \abno & \abno & \textbf{98.4} & 97.2 & 91.0 & 76.2 & 90.7 \\
2 & 3D & \abno & \abyes & \abno & 96.6 & 98.0 & 91.6 & 64.6 & 87.7 \\
3 & 4D & \abno & \abyes & \abno & 97.8 & 99.2 & \textbf{96.6} & 68.2 & 90.5 \\
\midrule
4 & \abno & \abyes & \abno & \abno & 97.8 & 93.2 & 88.8 & 72.8 & 88.2 \\
\rowcolor{ablrow}
5 & 4D & \abyes & \abyes & \abno & 94.6 & 98.2 & 79.0 & 63.0 & 83.7 \\
6 & 4D & \abyes & \abno & \abyes & 97.2 & 97.4 & 91.4 & 77.2 & 90.8 \\
\midrule
7 & 3D & \abyes & \abyes & \abyes & 97.2 & 98.6 & 81.0 & 63.2 & 85.0 \\
\rowcolor{oursrow}
8 & 4D & \abyes & \abyes & \abyes & \textbf{98.4} & \textbf{99.6} & 92.6 & \textbf{83.8} & \textbf{93.6} \\
\midrule
\multicolumn{10}{l}{\textit{(b) Temporal supervision:} $\mathcal{L}_{\mathrm{temp}}$ \textit{given as} $(w_{\mathrm{state}}, w_{\mathrm{change}}, w_{\mathrm{read}})$} \\
\addlinespace[1pt]
\rowcolor{ablrow}
5  & 4D & \abyes & \abyes & $(0,0,0)$ & 94.6 & 98.2 & 79.0 & 63.0 & 83.7 \\
9  & 4D & \abyes & \abyes & $(1,0,1)$ & 97.0 & \textbf{99.6} & \textbf{93.4} & 76.0 & 91.5 \\
10 & 4D & \abyes & \abyes & $(0,4,0)$ & 97.0 & \textbf{99.6} & 91.4 & \textbf{84.8} & 93.2 \\
11 & 4D & \abyes & \abyes & $(1,1,1)$ & 96.6 & 99.0 & 89.4 & 83.6 & 92.2 \\
\rowcolor{oursrow}
8  & 4D & \abyes & \abyes & $(1,4,1)$ & \textbf{98.4} & \textbf{99.6} & 92.6 & 83.8 & \textbf{93.6} \\
\bottomrule
\end{tabular}
}
\end{center}
\end{table}

We conduct controlled ablations on LIBERO to examine the history pathway, geometric targets, and temporal supervision, and to analyze what the learned history tokens encode and whether they contribute to action prediction.
All variants in Table~\ref{tab:ablation} are trained for 10k steps with a global batch size of 64 on four NVIDIA A800 GPUs, using the same backbone initialization, training data, optimizer, learning-rate schedule, and evaluation protocol.

\paragraph{Model Components.}
We first evaluate the contributions of the history pathway and alignment objectives in Table~\ref{tab:ablation}(a).
Adding the history pathway alone (row 4) reduces the average success rate from 90.7\% to 88.2\%, while current-frame alignment with 4D targets alone (row 3) achieves 90.5\%.
Combining the history pathway with both alignment objectives (row~8) raises the average to 93.6\%, with the largest gain on Long (83.8\% vs.\ 76.2\%). 
History inputs alone do not improve the baseline; the gain comes from pairing them with 4D representation supervision.

We then remove each alignment objective from the full model.
Without temporal alignment $\mathcal{L}_{\mathrm{temp}}$ (row~5),
the average success rate drops from 93.6\% to 83.7\%; removing
current-frame alignment $\mathcal{L}_{\mathrm{cur}}$ (row~6) reduces
it to 90.8\%. Both removals cause the largest degradation on Long,
from 83.8\% to 63.0\% and 77.2\%, respectively. Current-frame alignment
improves the history-augmented policy when paired with temporal
supervision, but degrades it without that supervision
(rows~4--6 and~8).
\begin{figure}[t]
\centering
\includegraphics[width=\linewidth]{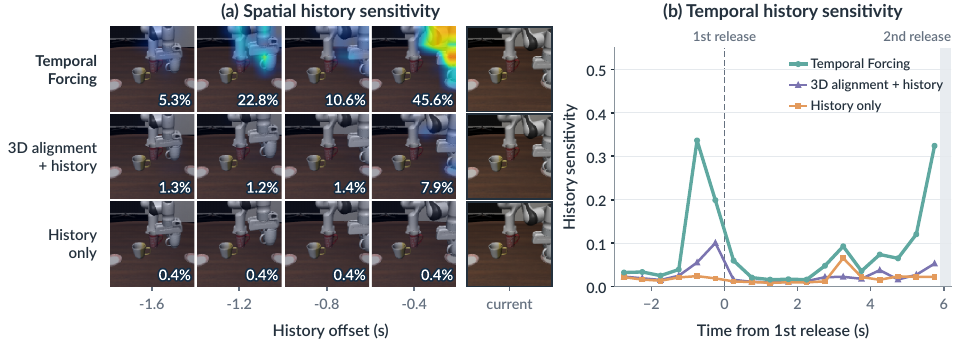}
\caption{\textbf{History utilization under controlled interventions.}
(a) Patch-level responses: warmer colors indicate larger relative action changes on a shared 0--50\% scale; numbers show the maximum response per frame.
(b) Relative responses summed over six separate frame interventions and averaged across six episodes aligned to the first release.
The dashed line marks the first release; shading begins at the earliest second release.}
\label{fig:probe}
\end{figure}
\paragraph{Geometric Targets.}
Context-conditioned 4D targets outperform framewise 3D targets under
both alignment configurations in Table~\ref{tab:ablation}(a). We obtain the 3D targets
by processing each frame independently with the same StreamVGGT
architecture and pretrained weights. With current-frame alignment
alone, 4D targets raise the average success rate from 87.7\% to 90.5\%
(rows~2 and~3). The advantage grows to 8.6 percentage points in the
full history-alignment model, from 85.0\% to 93.6\%
(rows~7 and~8).

\paragraph{Latent-Transition Supervision.}
We next examine the terms of $\mathcal{L}_{\mathrm{temp}}$ in Table~\ref{tab:ablation}(b), keeping the history pathway, 4D targets, and current-frame alignment fixed.
Removing the change loss while retaining state and readout supervision (row 9) reduces average success from 93.6\% to 91.5\% and Long success from 83.8\% to 76.0\%.
Change supervision alone (row 10) achieves 93.2\% average success and 84.8\% on Long, showing that matching temporal feature differences provides an effective training signal for the history pathway.
Equal weighting of the three terms (row 11) yields 92.2\% average success.
The default $(1,4,1)$ weighting achieves the highest average across suites, while change-only supervision performs best on Long.

\paragraph{History Utilization.}
\begin{wraptable}{r}{0.44\textwidth}
\centering
\small
\setlength{\tabcolsep}{4pt}
\renewcommand{\arraystretch}{1.05}
\caption{Linear probing of motion over the observed history interval
from history tokens on LIBERO-Long ($R^2$).}
\label{tab:hist_probe}
\vspace{4pt}
\adjustbox{max width=\linewidth}{%
\begin{tabular}{@{}l cc@{}}
\toprule
Model & End-effector & Object \\
\midrule
Untrained pathway & 0.48 & 0.74 \\
History only (row 4) & 0.44 & 0.78 \\
w/o $\mathcal{L}_{\mathrm{temp}}$ (row 5) & 0.51 & 0.64 \\
3D alignment (row 7) & 0.53 & 0.79 \\
\rowcolor{oursrow} \textbf{Temporal Forcing (row 8)} & \textbf{0.83} & \textbf{0.90} \\
\bottomrule
\end{tabular}}
\end{wraptable}
To examine what the learned history retains, we linearly probe the 16 history
tokens for end-effector and object displacement between the oldest and most
recent history frames, using simulator-state targets independent of the 4D
foundation model. Action supervision alone yields representations no more
predictive than an untrained pathway ($R^2$ 0.44 vs.\ 0.48), whereas Temporal
Forcing makes both displacements substantially more decodable ($R^2$ 0.83 and 0.90;
Table~\ref{tab:hist_probe}), showing that history access alone does not determine what the pathway retains.

We then close the injection gate at inference to measure the
contribution of the learned history pathway. For the main model,
this reduces the average success rate from 98.8\% to 97.2\%
(Table~\ref{tab:main}). Under the 10k-step protocol, closing the gate reduces
the average success rate by 5.4 percentage points for Temporal Forcing (row~8),
whereas it slightly increases that of the history-only variant (row~4), by 0.3 points.

We further examine where and when history affects action prediction
through interventions on past observations (Fig.~\ref{fig:probe}). Temporal Forcing
shows stronger patch-level responses than the history-only and
3D-alignment variants (Fig.~\ref{fig:probe}(a)). Averaged over six episodes aligned
to the first release, its response to frame-level interventions rises
before the first release, drops immediately afterward, and increases
again before the second release (Fig.~\ref{fig:probe}(b)). The influence of history
is thus stage-dependent, with stronger responses around the placement
transitions.

\subsection{Real-Robot Experiments}
\label{sec:real}
\begin{figure}[t]
\centering
\includegraphics[width=\linewidth]{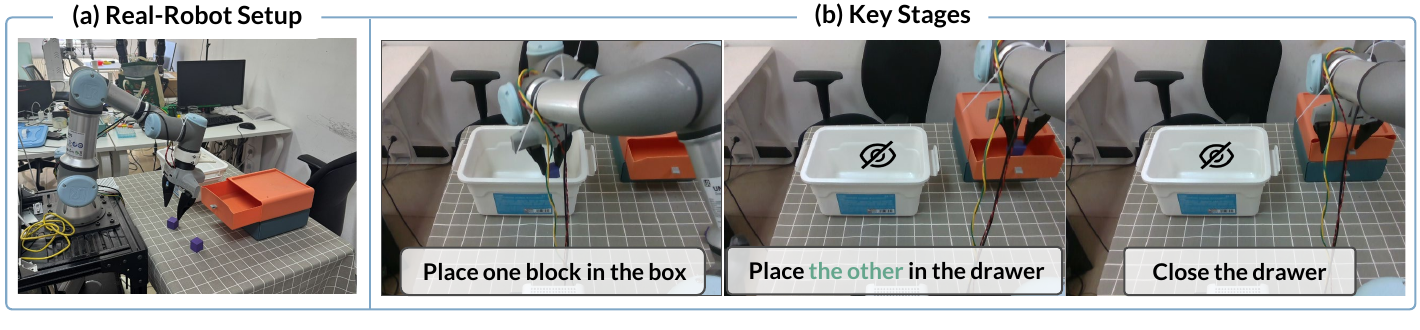}
\caption{\textbf{Real-robot hidden-placement task.}
(a) Experimental setup.
(b) Three sequential stages.
The crossed-out eye indicates that the first block is hidden in the box.}
\label{fig:real}
\end{figure}
We conduct real-robot experiments on a UR3 arm operating at 10\,Hz, with a static RealSense D435 and a wrist-mounted fisheye camera.
The task involves placing one of two identical blocks into an opaque box, placing the other into a drawer, and then closing the drawer (Fig.~\ref{fig:real}).
The first block becomes hidden after placement, making this a multi-stage setting in which earlier interactions are no longer fully visible as execution proceeds. 
All compared models are fine-tuned on the same 100 demonstration trajectories.
For each model, we evaluate each stage independently over 30 trials, using a stage-specific instruction and only the objects required for that stage.
We also evaluate the complete sequence over 30 trials without intervention and report cumulative stage success, where a stage counts as successful only when all preceding stages have also succeeded.

\begin{table}[t]
\caption{\textbf{Real-robot performance (successes/trials).}
Isolated stages are evaluated independently; sequential columns report cumulative success through the first, second, and third stages.
Best results in each column are in bold.}
\label{tab:real}
\begin{center}
\small
\setlength{\tabcolsep}{9pt}
\adjustbox{max width=\textwidth}{
\begin{tabular}{@{}l|cccc|ccc@{}}
\toprule
& \multicolumn{4}{c}{Isolated stages} & \multicolumn{3}{c}{Sequential task} \\
\cmidrule(lr){2-5}\cmidrule(lr){6-8}
\multicolumn{1}{c|}{Method} & Box & Drawer & Close & Total & $S_1$ & $S_{1:2}$ & $S_{1:3}$ \\
\midrule
OpenVLA & 24/30 & 13/30 & 21/30 & 58/90 & 18/30 & 5/30 & 0/30 \\
$\pi_0$ & 26/30 & 19/30 & 25/30 & 70/90 & 22/30 & 10/30 & 4/30 \\
StarVLA-OFT (base) & 28/30 & 21/30 & \textbf{26/30} & 75/90 & 25/30 & 12/30 & 6/30 \\
\rowcolor{oursrow} \textbf{Temporal Forcing (Ours)} & \textbf{29/30} & \textbf{24/30} & 25/30 & \textbf{78/90} & \textbf{27/30} & \textbf{20/30} & \textbf{13/30} \\
\bottomrule
\end{tabular}
}
\end{center}
\end{table}

As shown in Table~\ref{tab:real}, Temporal Forcing and the base model achieve similar success counts on the isolated stages (78/90 vs.\ 75/90) and the first stage of the sequence (27/30 vs.\ 25/30).
The performance gap becomes larger as sequential execution proceeds: Temporal Forcing completes the first two stages in 20/30 trials, compared with 12/30 for the base model.
Full-task success increases from 6/30 (20.0\%) to 13/30 (43.3\%).
This pattern shows that the gains of Temporal Forcing are more pronounced in sequential execution than on the constituent manipulation stages evaluated independently.
On a single NVIDIA GeForce RTX 5090 GPU, inference takes 58\,ms per action chunk for Temporal Forcing and 45\,ms for the base model, both within the 100\,ms control interval.

\section{Conclusion}

We presented Temporal Forcing, which explicitly supervises VLA history representations with causal 4D geometric targets. It aligns both latent temporal states and their transitions during training, while retaining only a compact History Pathway at inference. Across LIBERO, RoboTwin 2.0, and real-robot sequential manipulation, Temporal Forcing consistently improves performance, and controlled interventions confirm that the learned history directly contributes to action prediction. These results suggest that effective history-aware VLAs require not only access to past observations, but also explicit supervision of what their history representations retain.

\section*{AI use statement}
Large Language Models (LLMs) were used for grammar and wording polishing, literature search, and feedback on experimental design and result interpretation. All experiments were conducted by the authors, and all research content and contributions are entirely the responsibility of the authors.

\section*{Reproducibility Statement}

Sections~\ref{sec:method} and~\ref{sec:experiments}, and Appendix~\ref{app:implementation} provide the model, training, evaluation, and
4D-target construction details, while Appendix~\ref{app:analyses} documents the controlled
analyses and representation probes. Code and real-robot rollout videos will be released publicly.

\bibliography{iclr2027_conference}
\bibliographystyle{iclr2027_conference}
\newpage
\appendix

\section{Implementation and Training Details}
\label{app:implementation}

\paragraph{4D target construction.}
We use StreamVGGT to extract geometric targets offline from each training
trajectory in causal order. On LIBERO, the model processes causal segments of
up to 12 seconds. On RoboTwin and the real robot, it uses the episode prefix
available at each anchor. We extract layer-21 features every four timesteps
and store them in fp16. For temporal alignment, we spatially average the
feature grid at each historical timestep to obtain a pooled vector for each
camera, without normalization. For current-frame alignment, we average-pool
the $37\times37$ Dense Geometric Feature grid to $8\times8$.

We subtract a fixed per-camera mean $\mu_c$, estimated once from the training
set of each benchmark, from each pooled vector. The state and readout losses
use the layer-normalized centered vectors as targets, while the dense targets
are directly layer-normalized. The change loss uses differences between
adjacent centered vectors without further normalization, since normalizing
each timestep separately would distort their difference. We mask pairs whose
target difference has an $\ell_2$ norm below $10^{-4}$; these mainly occur
near the start of an episode, where the history window duplicates frames.

For the framewise 3D control in Table~\ref{tab:ablation}(a), we use the same StreamVGGT
architecture and pretrained weights but process each frame independently.
This comparison isolates the role of temporal context in the geometric
targets.

\paragraph{History Pathway.}
We first encode each past frame at $224\times224$ resolution using a frozen
DINOv2 ViT-L/14. A lightweight Q-Former then compresses its features into two
512-dimensional gist tokens. After adding time-offset and camera embeddings,
a two-layer causal Transformer aggregates the ordered token sequence.
Sixteen learned queries summarize these features into a fixed set of history
tokens, independent of the history-window length. The history tokens are
injected into the primary-camera image tokens before the first VLA decoder
layer through gated cross-attention. The gate is initialized to zero, so the
fusion block initially preserves the base model's output. Current-frame
alignment uses VLA layer 24 and StreamVGGT layer 21. During inference,
StreamVGGT and the alignment heads are removed, and per-frame gist tokens are
cached before adding time-offset embeddings for reuse across control steps.

\paragraph{Training settings.}
Across experiments, we use a learning rate of $10^{-5}$ for the Qwen3-VL
backbone and $10^{-4}$ for the action head, the History Pathway, and the
alignment heads. Benchmark-specific training, history, and evaluation settings are summarized
in Table~\ref{tab:training_config}.

On RoboTwin, the History Pathway and both alignment objectives use only the
head camera. The frame offsets are the same as on LIBERO and correspond to
3.7 seconds at 15 frames per second. For real-robot fine-tuning, we freeze
the backbone vision tower for both Temporal Forcing and the base model.
OpenVLA and $\pi_0$ are fine-tuned on the same demonstrations using their
official recipes.

\paragraph{Computational cost.}
The History Pathway adds 35.2M trainable parameters (about 0.7\% of the base
model) and a frozen DINOv2 encoder with 304.4M parameters; the alignment heads
(10.5M) are used only during training. Extracting the 4D targets is a one-time
preprocessing step that takes 10.9 GPU hours for LIBERO and produces a 68~GiB
fp16 cache.

\begin{table}[t]
\small
\caption{Training settings across benchmarks. Main and Controlled denote the
main LIBERO experiments and the 10k-step controlled studies, respectively.}
\label{tab:training_config}
\setlength{\tabcolsep}{5pt}
\begin{center}
\begin{tabular*}{\linewidth}{@{\extracolsep{\fill}}l c c c c@{}}
\toprule
 & \multicolumn{2}{c}{LIBERO} & RoboTwin & Real robot \\
\cmidrule(lr){2-3}
 & Main & Controlled & & \\
\midrule
Training steps & 50k & 10k & 10k & 8k \\
Global batch size & 128 & 64 & 64 & 48 \\
GPUs & 8$\times$A100 & 4$\times$A800 & 4$\times$A800 & 4$\times$A800 \\
Control frequency (Hz) & 20 & 20 & 15 & 10 \\
Window offsets (frames) & \multicolumn{2}{c}{$-56 \ldots -8$} & $-56 \ldots -8$ & $-28 \ldots -4$ \\
Window span (s) & 2.8 & 2.8 & 3.7 & 2.8 \\
History cameras & 2 & 2 & 1 (head) & 2 \\
Alignment cameras & 2 & 2 & 1 (head) & 2 \\
Anchor stride & 4 & 4 & 4 & 4 \\
Causal context & 12 s & 12 s & episode & episode \\
Action dim / horizon & 7 / 8 & 7 / 8 & 14 / 16 & 7 / 8 \\
\bottomrule
\end{tabular*}
\end{center}
\end{table}
\section{Additional Analyses}
\label{app:analyses}

\begin{figure}[t]
\centering
\includegraphics[width=\textwidth]{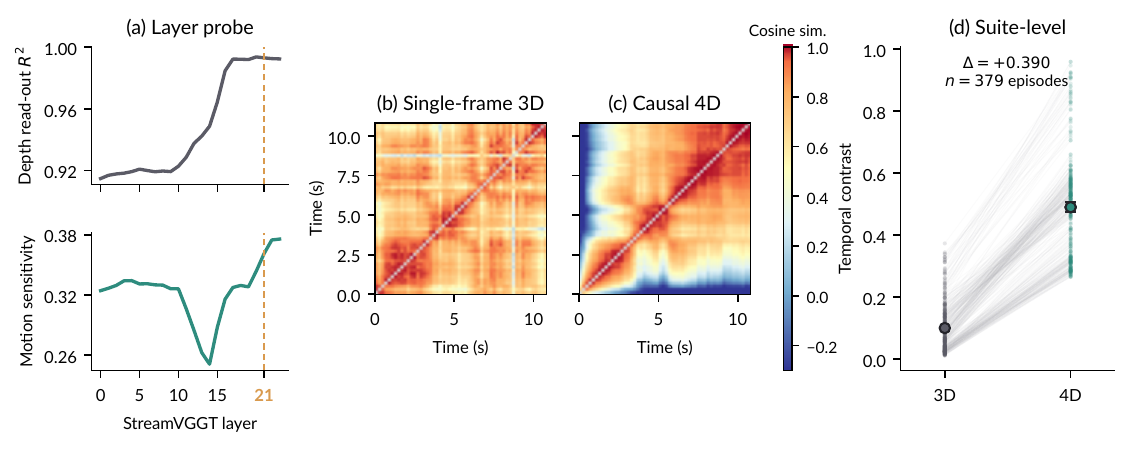}
\caption{Analysis of StreamVGGT features. (a) Depth readout and motion
sensitivity across layers. (b--c) Pairwise cosine similarity of framewise 3D
and causal 4D targets from the same trajectory, with diagonals masked.
(d) Temporal contrast on LIBERO-Long.}
\label{fig:probe4d3d}
\end{figure}

\paragraph{4D target layer selection.}
We first examine depth readout and motion sensitivity at different StreamVGGT
layers in Fig.~\ref{fig:probe4d3d}(a). We sample 120 anchor windows from
primary-camera trajectories in LIBERO-Spatial, LIBERO-Object, and LIBERO-Long.
Each probe window contains eight log-spaced frames at offsets
$[-64,-32,-16,-8,-4,-2,-1,0]$; this sampling is separate from the policy's
history window. For each layer, a ridge regressor with an $\ell_2$ coefficient
of 10 predicts patch-wise log-depth produced by the frozen StreamVGGT depth
head. We use 256 sampled current-frame patch features per anchor and split
anchors into 80\% training and 20\% test sets, reporting held-out $R^2$.
Motion sensitivity measures the Pearson correlation between the $\ell_2$
feature change over the final two frames and the corresponding patch-pooled
absolute pixel change. We compute this correlation across spatial patches
and average it over anchors.

Depth readout saturates in the late layers, while motion sensitivity continues
to vary. We select layer 21, the earliest layer to reach both depth
$R^2\geq0.99$ and motion correlation $\geq0.36$. The final two layers increase
motion correlation only modestly, to at most 0.376.

\paragraph{Temporal structure of geometric targets.}
Rows 2 and 7 of Table~\ref{tab:ablation}(a) replace the causal 4D targets with framewise 3D
targets and perform worse than their 4D counterparts. To examine this
difference, we compare the two targets from the same trajectory in
Fig.~\ref{fig:probe4d3d}(b--c). The 3D targets assign high similarity to many
temporally distant observations, whereas the causal 4D targets separate early
states from later ones. To quantify this difference, we define temporal
contrast as the mean cosine similarity between nearby frames, with a lag of
at most one eighth of a segment, minus that between distant frames, with a
lag of at least one third. Over the first 60-anchor (12-second) segment of
each of 379 LIBERO-Long episodes, temporal contrast increases from 0.101 for
3D targets to 0.491 for causal 4D targets (Fig.~\ref{fig:probe4d3d}(d)).
The causal targets thus distinguish states at different points in a
trajectory more clearly, whereas differences between framewise 3D targets
carry less information about how the scene has evolved.

\begin{figure}[t]
\centering
\includegraphics[width=\textwidth]{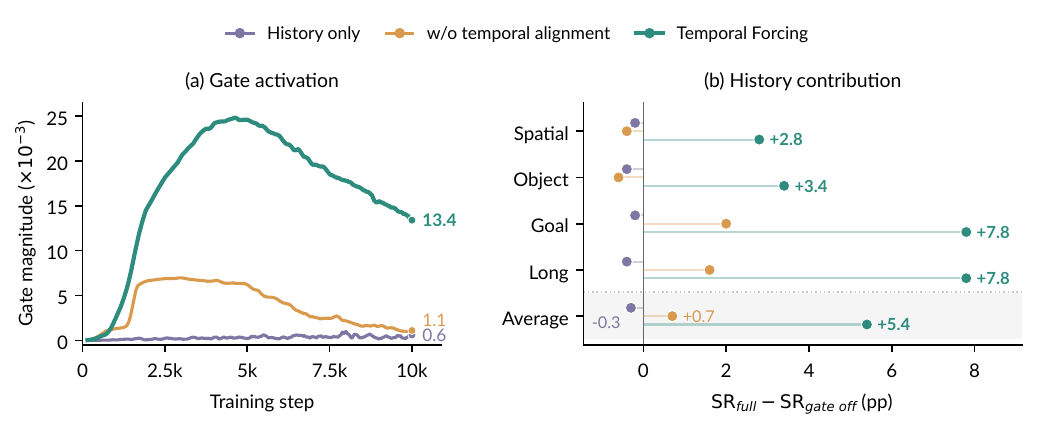}
\caption{History utilization under the 10k-step controlled protocol.
(a) Injection-gate magnitude during training. (b) Success-rate difference
between the original policy and the same policy with its gate closed at
inference.}
\label{fig:mechanism}
\end{figure}

\paragraph{Gate dynamics.}
Fig.~\ref{fig:mechanism}(a) shows how the injection gate develops during
training. With the action loss alone (row 4 of Table~\ref{tab:ablation}(a)), the gate
magnitude remains near $0.6\times10^{-3}$. Adding current-frame alignment
without temporal alignment (row 5) opens the gate temporarily, but its
magnitude then falls to $1.1\times10^{-3}$, indicating that the model learns
to down-weight the unsupervised pathway. With temporal alignment (row 8), the
gate reaches $13.4\times10^{-3}$, about 22 times the action-loss-only value,
and remains open.

\paragraph{History removal at inference.}
To evaluate the contribution of the History Pathway, we set the injection
gate to zero while keeping the current observation and model weights
unchanged. The gated fusion block then becomes an identity mapping.
We report the success-rate difference
$\Delta\mathrm{SR}=\mathrm{SR}_{\mathrm{full}}-\mathrm{SR}_{\mathrm{gate\ off}}$
in Fig.~\ref{fig:mechanism}(b). Under the 10k-step protocol, this difference
is $-0.3$ percentage points for the history-only model, whose success slightly
increases on all four suites when the gate is closed, and $+0.7$ for the
variant without temporal alignment. For Temporal Forcing, closing the gate
reduces average success by 5.4 percentage points, with the largest drops on
Goal and Long (7.8 points each). These results show that the aligned History
Pathway contributes to task success at inference, beyond its role during
training.
\paragraph{History-representation probes.}
\label{app:history-probe}

Table~\ref{tab:hist_probe} probes the 16 history tokens $M_t$ produced by the
models in Table~\ref{tab:ablation} under the 10k-step protocol. Each checkpoint
is frozen and evaluated with the same history window used by the policy
(offsets $-56$ to $-8$ frames, both cameras). We record $M_t$ every eighth
timestep after a full history window is available and flatten the 16 tokens into
an 8{,}192-d feature. From the ten LIBERO-Long tasks, we use 12 demonstrations
per task and split the resulting episodes into 70, 20, and 30 episodes for
probe training, validation, and testing, respectively. The split is shared
across all models.

The probe targets measure motion between the oldest and most recent observed
history frames. The end-effector target is the displacement magnitude
$\|\mathbf{p}_{t-8}-\mathbf{p}_{t-56}\|_2$, computed from recorded
proprioception. The object target is the maximum displacement magnitude over
all task objects,
$\max_j \|\mathbf{q}^{j}_{t-8}-\mathbf{q}^{j}_{t-56}\|_2$,
where object positions are recovered from the recorded simulator states by
loading each state into the LIBERO environment.

We fit ridge regressors to standardized, PCA-reduced 256-d features, with both
standardization and PCA fitted on the probe-training episodes only. The
regularization strength is selected from
$\{0.01,0.1,1,10,100\}$ using validation $R^2$, and test $R^2$ is reported.
The untrained-pathway control reinitializes the Q-Former, temporal Transformer,
and output queries of the Temporal Forcing checkpoint, yielding history tokens
produced by randomly initialized temporal modules on top of frozen DINOv2
features.

\paragraph{Interventions on past observations.}
For Fig.~\ref{fig:probe}, each intervention replaces part of one input frame
with the content 0.4~s earlier (0.4~s later for the oldest slot) and keeps all
other inputs and weights unchanged. The response is the relative change of the
predicted action chunk in the normalized action space,
$r=\|\hat{\mathbf{A}}'_t-\hat{\mathbf{A}}_t\|_F/\|\hat{\mathbf{A}}_t\|_F$.
In the frame-level analysis (Fig.~\ref{fig:probe}(b)), both camera images of
one history slot are replaced, and responses are summed over the six slots
from $-2.8$ to $-0.8$~s and averaged over six episodes. In the patch-level
analysis (Fig.~\ref{fig:probe}(a)), each of the $7\times7$ patches of the
primary-camera image is replaced in turn, for each history slot and the
current frame, on a single episode; the figure shows the four most recent
slots.

\begin{table}[t]
\small
\caption{History-window ablation on LIBERO (success rate, \%). Every row
uses the full Temporal Forcing recipe of Table~\ref{tab:ablation} (row~8)
under the 10k-step protocol and changes only the history window: $K$ past
frames per camera at spacing $\Delta t$, spanning the stated interval before
the current step (LIBERO runs at 20 frames per second). The teal row is the
default configuration. Best results in each column are in bold.}
\label{tab:history_window}
\setlength{\tabcolsep}{6pt}
\renewcommand{\arraystretch}{1.06}
\begin{center}
\begin{tabular}{c c c cccc c}
\toprule
\multicolumn{3}{c}{History window} & \multicolumn{4}{c}{LIBERO suite} & \\
\cmidrule(lr){1-3}\cmidrule(lr){4-7}
Window (s) & $K$ & $\Delta t$ (s) & Spatial & Object & Goal & Long & Average \\
\midrule
$[-1.4,\,-0.2]$  & 7  & 0.2 & 88.6 & 87.6 & 84.2 & 53.8 & 78.6 \\
$[-2.8,\,-0.2]$  & 14 & 0.2 & \textbf{99.0} & 99.4 & 93.4 & 66.6 & 89.6 \\
\rowcolor{oursrow}
$[-2.8,\,-0.4]$  & 7  & 0.4 & 98.4 & \textbf{99.6} & 92.6 & \textbf{83.8} & \textbf{93.6} \\
$[-3.2,\,-0.8]$  & 4  & 0.8 & 98.6 & \textbf{99.6} & 83.2 & 79.4 & 90.2 \\
$[-5.6,\,-0.8]$  & 7  & 0.8 & 97.2 & 98.6 & \textbf{96.8} & 81.8 & \textbf{93.6} \\
$[-11.2,\,-1.6]$ & 7  & 1.6 & 95.4 & 99.2 & 96.6 & 72.2 & 90.9 \\
\bottomrule
\end{tabular}
\end{center}
\end{table}

\paragraph{History window.}
With the model components and geometric targets fixed, we vary the
temporal coverage and sampling density of the history window in
Table~\ref{tab:history_window}. With $K=7$ frames, increasing the maximum
lookback from 1.4~s to 2.8~s raises the average success rate from 78.6\% to
93.6\%. Extending it to 5.6~s maintains the same average, while 11.2~s
reduces it to 90.9\%. At the 2.8~s horizon, halving the sampling interval to
0.2~s (14 frames) reduces the average to 89.6\% and Long success from 83.8\%
to 66.6\%, and doubling it to 0.8~s (4 frames, 3.2~s lookback) reduces the average to
90.2\%, mainly on Goal (83.2\%). Longer or denser history is therefore not
uniformly better. We use seven frames at 0.4~s intervals, which match the
best average performance and achieve the highest success rate on Long.

\section{Additional Results and Visualizations}
\label{app:results}

\begin{table}[t]
\small
\caption{Complete RoboTwin~2.0 success rates (\%) in the easy (clean)
configuration. Temporal Forcing is evaluated over
100 episodes per task. Best results in each row are in bold.}
\label{tab:robotwin_full}
\setlength{\tabcolsep}{10pt}
\renewcommand{\arraystretch}{1.08}
\begin{center}
\begin{tabular}{@{}l c c c c@{}}
\toprule
Task & $\pi_0$ & DP3 & StarVLA-OFT & Ours \\
\midrule
Click Bell & 44 & 90 & 94 & \textbf{100} \\
Place Burger Fries & 80 & 72 & \textbf{96} & 92 \\
Place Cans Plasticbox & 34 & 48 & 81 & \textbf{88} \\
Stack Blocks Two & 42 & 24 & 83 & \textbf{86} \\
Handover Mic & 98 & \textbf{100} & 39 & 79 \\
Place Bread Skillet & 23 & 19 & 56 & \textbf{59} \\
Place Bread Basket & 17 & 26 & 52 & \textbf{58} \\
Blocks Ranking RGB & 19 & 3 & 45 & \textbf{57} \\
Handover Block & 45 & \textbf{70} & 0 & 44 \\
Stack Blocks Three & 17 & 1 & \textbf{41} & 40 \\
Blocks Ranking Size & 7 & 2 & 27 & \textbf{29} \\
Place Dual Shoes & 15 & 13 & \textbf{28} & 22 \\
\midrule
Average & 36.8 & 39.0 & 53.5 & \textbf{62.8} \\
\bottomrule
\end{tabular}
\end{center}
\end{table}
\paragraph{Complete RoboTwin results.}
Table~\ref{tab:robotwin_full} reports results on all twelve evaluated tasks.
Temporal Forcing improves over the base model on nine tasks, increasing
average success from 53.5\% to 62.8\%. The largest gains are on Handover
Block and Handover Mic, from 0\% to 44\% and from 39\% to 79\%, respectively.
Both tasks involve object occlusion during transfer between the arms. The
three regressions occur on tasks in which completed stages remain visible.

\paragraph{Zero-shot robustness on LIBERO-Plus.}
LIBERO-Plus tests static distribution shifts rather than history-dependent
decisions, so we use it to check that the History Pathway does not reduce
robustness. Without additional training, Temporal Forcing improves pooled
success from 75.0\% to 77.8\% (Table~\ref{tab:additional_results}). It
improves six of the seven dimensions and matches the base model on lighting.
The largest gain occurs under sensor noise, where success increases from
73.1\% to 81.3\%.

\begin{table}[t]
\small
\caption{Zero-shot LIBERO-Plus success rates (\%). All models are trained
on standard LIBERO only. Total pools results over 10{,}030 perturbed tasks,
rather than averaging the seven dimensions equally. Baseline results for OpenVLA~\citep{openvla}, OpenVLA-OFT~\citep{openvlaoft},
$\pi_0$~\citep{pi0}, and $\pi_0$-Fast~\citep{pi0fast} are taken from
LIBERO-Plus~\citep{liberoplus}. Parentheses report gains over StarVLA-OFT in
percentage points. Best results in each column are in bold.}
\label{tab:additional_results}

\setlength{\tabcolsep}{3pt}
\renewcommand{\arraystretch}{1.05}

\begin{center}
\begin{tabular*}{\textwidth}{@{\extracolsep{\fill}}l c c c c c c c c@{}}
\toprule
Method & Camera & Robot & Language & Light & Background & Noise & Layout & Total \\
\midrule
OpenVLA
& 0.8 & 3.5 & 23.0 & 8.1 & 34.8 & 15.2 & 28.5 & 15.6 \\

OpenVLA-OFT
& 56.4 & 31.9 & 79.5 & 88.7 & 93.3 & 75.8 & 74.2 & 69.6 \\

$\pi_0$
& 13.8 & 6.0 & 58.8 & 85.0 & 81.4 & 79.0 & 68.9 & 53.6 \\

$\pi_0$-Fast
& \textbf{65.1} & 21.6 & 61.0 & 73.2 & 73.2 & 74.4 & 68.8 & 61.6 \\

\midrule
StarVLA-OFT (base)
& 47.0 & 60.1 & 87.0 & \textbf{96.3} & 95.3 & 73.1 & 79.2 & 75.0 \\

\addlinespace[2pt]
\textbf{Temporal Forcing}
& \shortstack{47.2\\[-1pt]\scriptsize(+0.2)}
& \shortstack{\textbf{63.7}\\[-1pt]\scriptsize(+3.6)}
& \shortstack{\textbf{88.7}\\[-1pt]\scriptsize(+1.7)}
& \shortstack{\textbf{96.3}\\[-1pt]\scriptsize(+0.0)}
& \shortstack{\textbf{97.4}\\[-1pt]\scriptsize(+2.1)}
& \shortstack{\textbf{81.3}\\[-1pt]\scriptsize(+8.2)}
& \shortstack{\textbf{81.6}\\[-1pt]\scriptsize(+2.4)}
& \shortstack{\textbf{77.8}\\[-1pt]\scriptsize(+2.8)} \\
\bottomrule
\end{tabular*}
\end{center}
\end{table}
\begin{figure}[t]
\centering
\includegraphics[width=\textwidth]{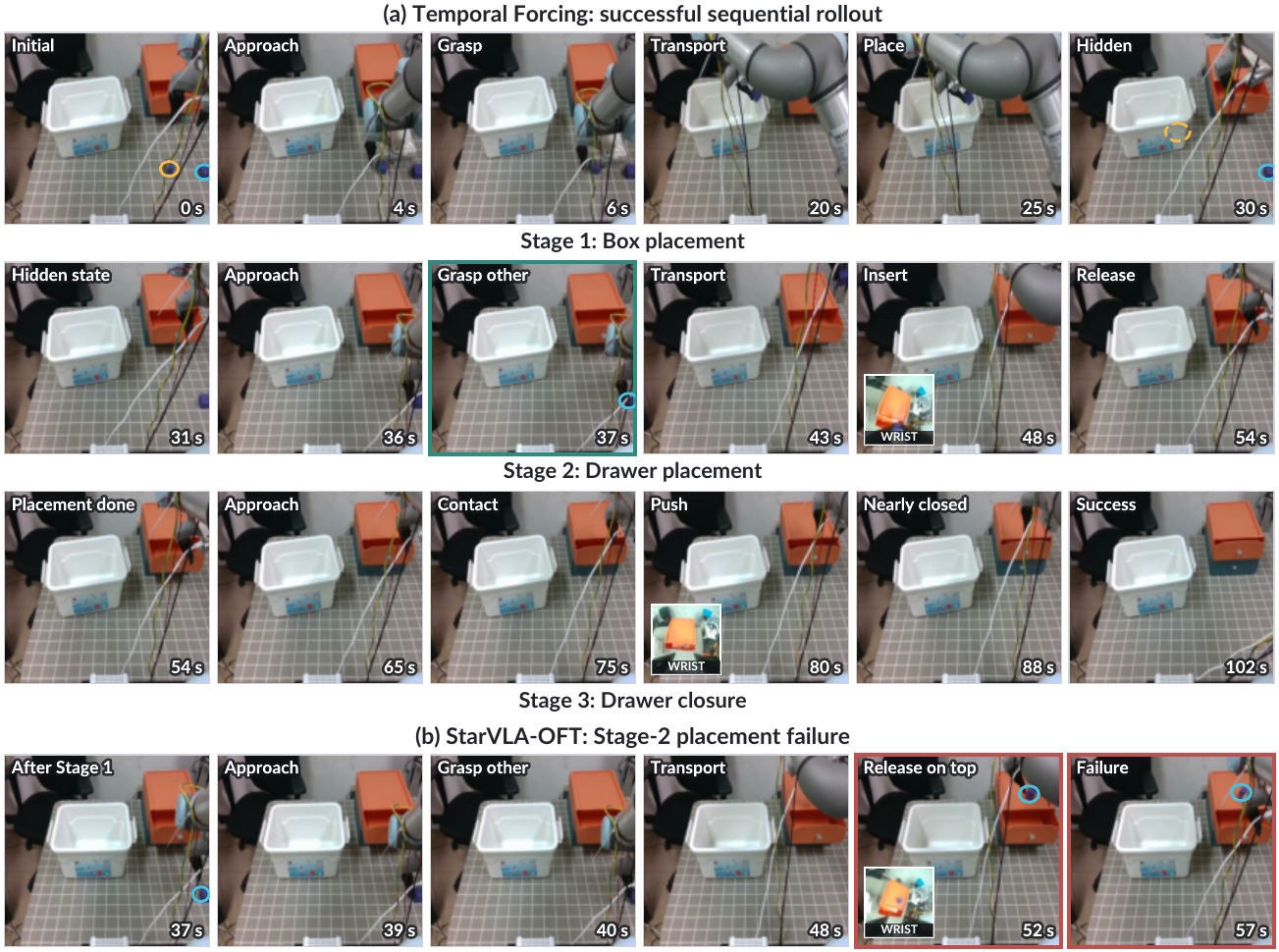}
\caption{Sequential real-robot rollouts. (a) Temporal Forcing completes all
three stages without a reset. (b) StarVLA-OFT releases the second block on
top of the drawer. Green and red borders mark key decision and failure
states. Circles identify the two blocks, and wrist-camera insets show
occluded manipulation details. All annotations are for visualization only.}
\label{fig:real_rollouts}
\end{figure}

\paragraph{Qualitative sequential rollouts.}
Fig.~\ref{fig:real_rollouts} compares complete execution sequences. After the first block is placed into the box, it is no longer visible, and the task proceeds to the remaining placement and drawer-closing stages. Temporal
Forcing selects the remaining block, places it into the drawer, and closes the
drawer without a reset. StarVLA-OFT also selects and transports the correct
block, but releases it on top of the drawer during the second stage.
Across the real-robot trials, Temporal Forcing and the base model complete
78/90 and 75/90 isolated stages, respectively, while their full-task successes
are 13/30 and 6/30. Together with the isolated-stage results, these rollouts illustrate that the larger performance difference emerges during continuous multi-stage execution rather than from a large difference in individual manipulation skills.

\end{document}